# IMAGE SEGMENTATION BY OPTIMAL AND HIERARCHICAL PIECEWISE CONSTANT APPROXIMATIONS

## M. Kharinov[1]

[1] St. Petersburg Institute for Informatics and Automation of RAS,
14–th line, 39, 199178, St. Petersburg, Russia
khar@iias.spb.su

Piecewise constant image approximations of sequential number of segments or clusters of disconnected pixels are treated. The method of majorizing of optimal approximation sequence by hierarchical sequence of image approximations is proposed. A generalization for multidimensional case of color and multispectral images is foreseen.

## Introduction

In terms of deductive Descriptive approach [3, 7] the preliminary stage of image transformation from the original form into a recognizable form is considered in the report. For this purpose, the piecewise constant image approximations are studied. According to the formal quality assessment the optimal approximations are the best, since they minimally differs from the image in the values of total squared error $E$ or standard deviation $\sigma$ of the image pixels from the averaged pixels of approximation. On the other hand, hierarchical approximations, compared with nonhierarchical ones, are far preferable for image recognition tasks. Whether these requirements are compatible - that is the question to be discussed.

## Analytical description

Let's consider the general case of multi-dimensional pixel clustering for color or multispectral images. Let $I_1$ and $I_2$ be two central data points of averages intensities for clusters 1 and 2, respectively. Let $n_1$ be the number of pixels in the cluster 1 and $n_2$ be the number of pixels in the cluster 2. Then the increment $\Delta E_{merge}$ of the total squared error $E$ caused by the merging of specified clusters and the reduction of the number of clusters per unit is given by the following formula:

$$\Delta E_{merge} = \frac{\|I_1 - I_2\|^2}{\frac{1}{n_1} + \frac{1}{n_2}} \geq 0 , \qquad (1)$$

where the symbol $\| \ \|$ denotes an Euclidean distance.

Just this quantity is minimized in the version [1] of Mumford-Shah model [10, 11]. In the version [6], the formula differs by an additive term, and in FLSA version [2] by a multiplicative factor to take into account the total length of the boundaries between the segments (clusters of connected pixels).

Let's write down the formula for splitting of the cluster 1, when its $k < n_1$ pixels with the central data point $I$, initiate a new cluster. In this case, cluster 1 is split into two clusters of $k$ and complementary $n_1 - k$ pixels, and cluster splitting is accompanied by increase of the cluster number per unit along with a non-positive increment $\Delta E_{split}$ of the total squared error:

$$\Delta E_{split} = -\frac{\|I - I_1\|^2}{\frac{1}{k} - \frac{1}{n_1}} \leq 0 . \qquad (2)$$

The composition of splitting and merging of clusters induces a correction operation without changing the number of clusters, which is accompanied by an increment $\Delta E_{correct}$ of the total squared error:

$$\Delta E_{correct} = \frac{\|I - I_2\|^2}{\frac{1}{k} + \frac{1}{n_2}} - \frac{\|I - I_1\|^2}{\frac{1}{k} - \frac{1}{n_1}}, \quad (3)$$

where the negative term in (3) describes the increment of the total squared error $E$, caused by converting of $k$ pixels from cluster 1 into a separate cluster, and the first term in (3) describes the increment of $E$ caused by merging of the initiated cluster with the cluster 2, in accordance with (1) and (2).

Noteworthy that by simplifying of the formula (3) $K$-means method [8, 9, 14] is derived. Applying (3) precisely, we have proposed for the clustering of pixel sets a more accurate method [5], which in one-dimensional case provides a calculation of a complete sequence of optimal image approximations that are treated in multi-threshold Otsu method [12, 13]. Then, using (1) and (2) we have developed top-down/bottom-up segmentation algorithm $F_u v$ that for given image $u$ and any masking image $v$ produces a hierarchical sequence $\{v_i\}$ of approximations $v_1, v_2, ..., v_g$. The sequence of approximations $F_u v = \{v_i\}$ depending on successive cluster number $i$ is described by a *convex* monotone sequence of corresponding values $E_1 \geq E_2 \geq ..., \geq E_g = 0$ of total squared error $E$:

$$E_i \leq \frac{E_{i-1} + E_{i+1}}{2}, \quad i = 2, 3, ..., g-1, \quad (4)$$

where $E_1$ corresponds to a trivial image approximation $v_1$, containing a single cluster, and $E_g = 0$ describes the last approximation $v_g = u$, containing the clusters of equal pixels. In the case of $v = u$ the algorithm $F_u u$ produces a sequence $\{u_i\}$ of approximations $u_1, u_2, ..., u_g$, which is reproduced when replacing the masking image by any of specified approximations:

$$F_u u_1 = F_u u_2 = ... = F_u u_g \equiv F_u u. \quad (5)$$

The property (5) is characteristic for the sequence of approximations $u_1, u_2, ..., u_g$, since in a different choice of image-mask, particularly, in the choice of an optimal approximation as masking image, this property is lost.

Thus, in one-dimensional case of gray-scale image, a sequence of optimal approximations is majorized by the self-consistent sequence of hierarchical approximations, which eventually should be generalized to the multidimensional case of color and multispectral images.

### Experimental results

Optimal and hierarchical approximations for standard Lenna image in 2, 4, 8 and 16 gradations are illustrated top-down in Fig. 1.

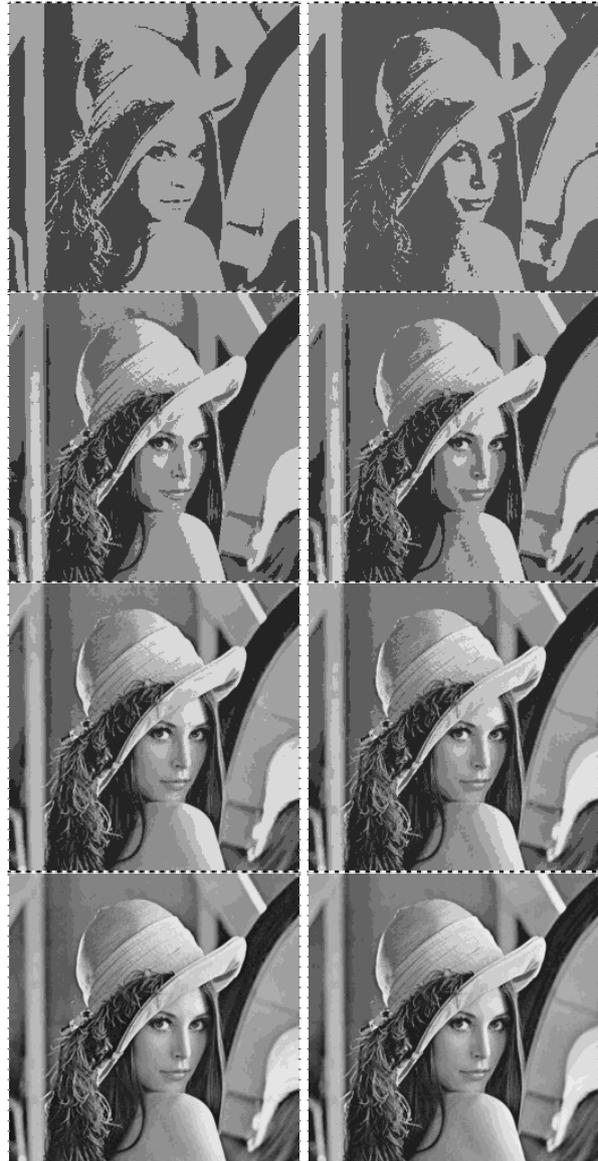

Fig. 1. Optimal and hierarchical image approximations.

Fig. 1 shows the optimal approximations on the left. The hierarchical approximations are shown on the right. Two bottom approximations visually coincide with each other and with the original image.

The sequential values $E_i$ of the total squared error for approximations containing up to 10 clusters of pixels are given in the table 1 and graphically illustrated in Fig. 2.

**Table 1. Total squared error**

| $g$ | Optimal | Hierarchical |
|---|---|---|
| 1 | 204664605,4 | 204664605,4 |
| 2 | 61548497,96 | 70364664,51 |
| 3 | 29502852,42 | 31708474,68 |
| 4 | 14675887,34 | 15629646,62 |
| 5 | 8967579,334 | 10149880,78 |
| 6 | 6605810,961 | 7647674,675 |
| 7 | 4691315,544 | 5568946,159 |
| 8 | 3697423,421 | 4054979,449 |
| 9 | 3042513,759 | 3447078,552 |
| 10 | 2473873,467 | 2853095,235 |

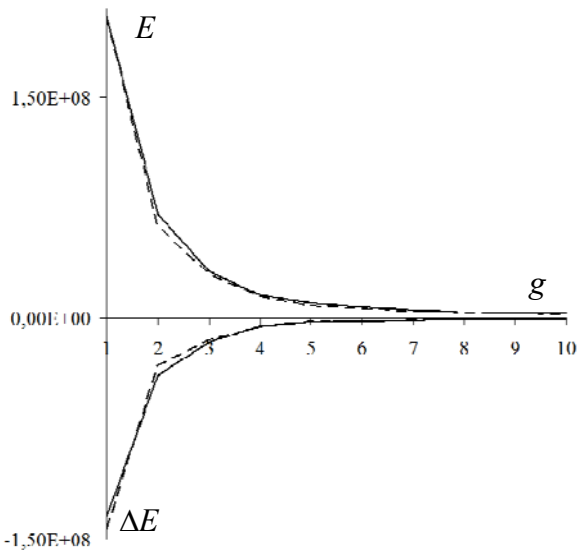

Fig. 2. Convex dependencies of $E$ on cluster number $g$.

Fig. 2 shows the behavior of the total squared error for the optimal (dashed line) and hierarchical (solid line) approximations of the image depending on the number of clusters. The upper graph shows a monotonic decrease of the total squared error $E_i$ itself and the lower graph shows the monotonic increase of its increment: $\Delta E_i = E_{i+1} - E_i, i = 1, 2, ..., g-1$. So, both dependencies are convex.

The overall results on segmentation are briefly presented graphically in Fig. 3, which demonstrates the dependencies of the standard deviation $\sigma$ on the cluster number in the range from 1 to 1000 ($\sigma$ is related with the total squared error by the formula: $E = N\sigma^2$).

In Fig. 2 the central gray curve separates the upper two curves describing partitioning of the image into connected segments from the lower curves that describe the image segmentation by clustering of disconnected pixels.

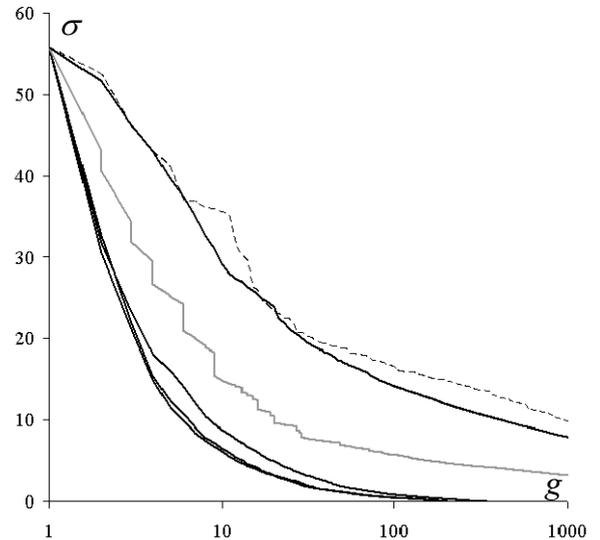

Fig. 3. Dependencies of $\sigma$ on cluster number $g$ (logarithmic scale along horizontal axes).

Upper dashed curve describes image segmentation according FLSA version [2] of Mumford-Shah model. The next curve just bellow dashed one describes segmentation according [1].

The gray curve demonstrates the calculations in our extended version of Mumford-Shah model, wherein the number of operations with the pixel sets is tripled according to (1)-(3). The algorithm is characterized as an algorithm of region growing, wherein just only the minimization of a total squared error determines how segment appears, how it is subdivided into the parts and how one segment captures the part of another with possible splitting of segment into disconnected pieces. The smooth curve sections correspond to hierarchical subsequences of image approximations and stepwise drops describe the conversion of one hierarchy into another.

Two of the lowest coalescing curves in Fig.3 describe the same optimal and majorizing approximations that illustrated by Fig. 2. The next overlying curve corresponds to a sequence of approximations generated by the algorithm $F_u v$ for the mask $v$ obtained by resizing of the image $u$ in sixteen times for each dimension. At first, the algorithm $F_u v$, starting from given approximation $v$ generates a complete sequence of approximations by top-down and bottom-up segmentation, and then edits them, converting into resulting approximation sequence $\{v_i\}$ that corresponds to a convex sequence $\{E_i\}$ (4). To apply the algorithm in multidimensional case of color and multispectral images, additional masking images should be introduced.

The image partitions for an increasing number of clusters do not depend from the linear transformation of intensities, and the algorithms of generating of image approximations commute with an image enlargement by doubling of pixels.

## Conclusion

Thus, in the report we perform a comparative analysis of segmentation techniques based on advanced *K*-means method, a complete multi-thresholding method according N. Otsu and also extended Mumford-Shah model with the triple number of operations with clusters or segments [4, 5].

We draw attention that a sequence of optimal approximations of an image in general case is not hierarchical. However, it turns out that a sequence of optimal approximations is majorized by a sequence of hierarchical approximations. Furthermore, in both cases the distances between the image and its approximations depending on a number of clusters are described by a convex sequence in values of total squared error that fall down with increasing number of clusters. It is equally important that our MOAS method of majorizing of optimal approximation sequence is accessible for generalization to the case of multidimensional case of colors and multispectral images.

In contrast to the nonhierarchical approximations, the hierarchical approximations provide fast generation, storage in a fixed lesser volume of RAM, as well as effective image processing, skipping the repetitions of clusters or segments. We therefore look forward to the implementation of our approach in image processing domain, especially in the practice of automatic object detection.